\documentclass[letterpaper, 10 pt, conference]{ieeeconf}
\IEEEoverridecommandlockouts 
\usepackage[l2tabu,orthodox]{nag}

\usepackage[
backend=biber, 
bibencoding=utf8,
style=ieee, 
sorting=none, 
natbib=true, 
doi=false, 
isbn=false, 
url=false, 
eprint=false, 
maxcitenames=1, 
mincitenames=1,
maxbibnames=5,
]{biblatex}

\usepackage[pdftex,colorlinks]{hyperref}
\usepackage[printonlyused]{acronym}

\usepackage{siunitx}
\usepackage[all]{nowidow}

\usepackage[dvipsnames]{xcolor}

\usepackage{lipsum}

\usepackage[pdftex]{graphicx}

\usepackage{epstopdf}

\usepackage{import}

\graphicspath{{./latexGoodPractices/}}

\usepackage{booktabs}

\usepackage{tabu}

\usepackage{amssymb,amsfonts,amsmath,amscd}

\usepackage{bm}

\newcommand{\bbm}{\begin{bmatrix}}
\newcommand{\ebm}{\end{bmatrix}}

\graphicspath{{Images/}}

\usepackage{subcaption}
\usepackage{float}
\usepackage{tabularx}
\usepackage{booktabs}

\usepackage{microtype}
\usepackage{csquotes}
\usepackage{multirow}
\usepackage{tabularx,booktabs}
\usepackage{array}
\usepackage{stfloats}
\usepackage{makecell}

\usepackage{authblk}

\title{\LARGE \textbf {%
Toward a Better Understanding of Robot Energy Consumption in Agroecological Applications 
}}

\author[1]{Alexis Bras$^{1}$, Alix Montanaro$^{1}$, Cyrille Pierre$^{1}$, Marilys Pradel$^{1}$ and Johann Laconte$^{1}$%

\thanks{$^{1}$ Université Clermont Auvergne, INRAE, UR TSCF, 63000, Clermont-Ferrand, France; name.surname@inrae.fr}%
}

\usepackage{xcolor}

\begin{document}

\maketitle

\thispagestyle{empty}
\pagestyle{empty}

\begin{abstract}

In this paper, we present a comprehensive analysis and discussion of energy consumption in agricultural robots. 
Robots are emerging as a promising solution to address food production and agroecological challenges, offering potential reductions in chemical use and the ability to perform strenuous tasks beyond human capabilities. 
The automation of agricultural tasks introduces a previously unattainable level of complexity, enabling robots to optimize trajectories, control laws, and overall task planning. 
Consequently, automation can lead to higher levels of energy optimization in agricultural tasks.
However, the energy consumption of robotic platforms is not fully understood, and a deeper analysis of contributing factors is essential to optimize energy use. 
We analyze the energy data of an automated agricultural tractor performing tasks throughout the year, revealing nontrivial correlations between the robot's velocity, the type of task performed, and energy consumption. 
This suggests a tradeoff between task efficiency, time to completion, and energy expenditure that can be harnessed to improve the energy efficiency of robotic agricultural operations.

\end{abstract}

\section{INTRODUCTION}
\label{sec:introduction}

As the global population continues to grow, the demand for food increases, placing significant pressure on agricultural systems. 
At the same time, climate change is reducing essential resources, further straining these systems. 
Agriculture is responsible for several environmental impacts among which are soil, water and air quality deterioration, soil erosion, water scarcity, and biodiversity erosion to name a few. Despite these serious negative impacts, agriculture can also be positive for the environment by trapping greenhouse gases within crops and soils, maintaining landscapes or mitigating climate change and biodiversity erosion through the adoption of certain farming practices \cite{OECD_2023}.
Therefore, agriculture plays a key role in environmental sustainability by sequestering carbon and improving soil health. 
Consequently, efficient and sustainable crop production is essential to meet the challenges posed by increasing environmental constraints.

\begin{figure}[t!]
    \centering
    \includegraphics[width=\linewidth]{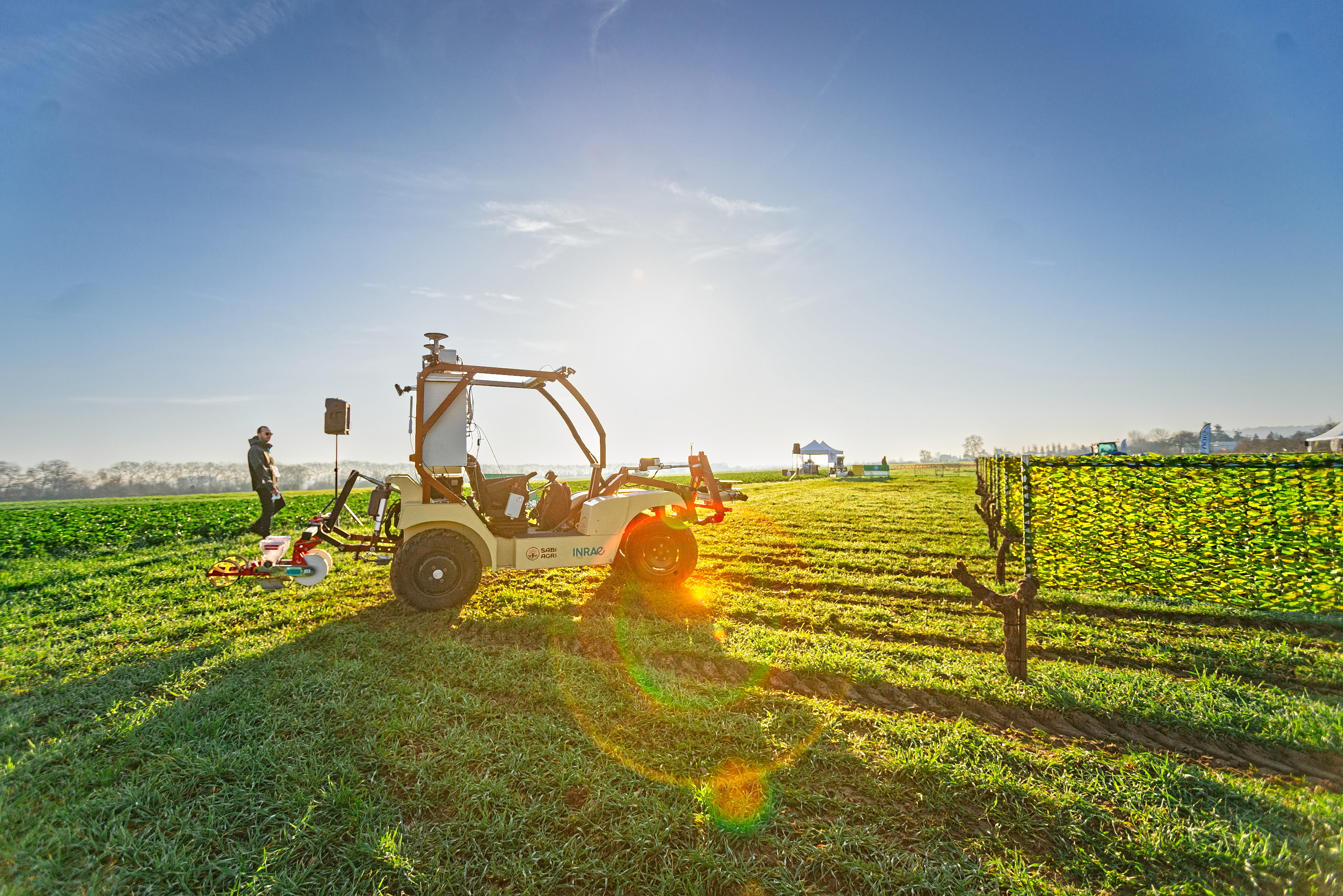}
    \caption{Example of an autonomous robot performing an agricultural task in a vineyard}
    \label{fig:intro}
\end{figure}

In this context, robotics has emerged as a promising solution to enhance the efficiency and productivity of agricultural practices \cite{lenain_agricultural_2021}. 
As seen in \autoref{fig:intro}, robots can perform repetitive and labor-intensive tasks with high precision, reducing the need for human labor and minimizing the waste of pesticides, seeds, and fertilizers. By taking over strenuous tasks, robots can improve working conditions for farmers and optimize resource use, leading to more sustainable agricultural practices.

However, integrating robotics into agriculture is not without its challenges. The environmental impact of electronic components, reliance on critical mineral resources, and high energy consumption raise sustainability concerns. 
Energy consumption in agricultural robotics is a significant concern because the source and efficiency of the energy used directly influence the robots' environmental impacts and overall sustainability. 
High energy consumption not only leads to greater environmental degradation but also increases operational costs, which can be a barrier to widespread adoption of these technologies. 
Therefore, optimizing energy use is essential to ensure that the benefits of robotic technology, such as increased efficiency and reduced labor costs, are not offset by negative environmental consequences. 
The key contributions of this paper are
\begin{itemize}
    \item An empirical study on the energy use of a robot performing agricultural tasks, offering valuable insights into real-world applications; and
    \item A discussion about practices for conducting experimental analyses of robotic energy consumption in complex outdoor environments.
\end{itemize} 

The remainder of this paper is organized as follows: Section \ref{sec:related_work} reviews existing literature on energy consumption in agriculture, global robotics, and specifically agricultural robotics. Section \ref{sec:empirical_analysis} presents an empirical study on the energy consumption of a robot performing agricultural tasks, detailing the experimental setup, tasks performed, and the environment in which the experiments were conducted. Section \ref{sec:discussion} discusses the results of the empirical study, including the impact of different speeds and tasks on energy consumption, and provides advice for monitoring energy consumption in complex outdoor environments. Finally, Section \ref{sec:conclusion} concludes the paper with a summary of findings and suggestions for future research directions.

\section{RELATED WORK}
\label{sec:related_work}

This section reviews existing literature on energy consumption in agricultural robotics, starting with energy usage in traditional agricultural machinery, followed by energy consumption in robotics more generally, and concluding with agricultural robots.

\subsection{Agriculture and Energy Consumption}

Agricultural machines are essential in modern farming practices due to their ability to significantly increase productivity and compensate for labor shortages \cite{lenain_agricultural_2021}. These machines, however, consume a considerable amount of energy, primarily from fossil fuels. As \citet{zegada-lizarazu_critical_2010} highlighted, energy inputs in agriculture are dominated by fuel use and the production of inputs such as fertilizers. Diesel-powered machinery accounts for roughly 30\% of total energy consumption, while fertilizers can contribute up to 70\% in certain crop systems. Field operations such as plowing, seeding, fertilizing, and harvesting are among the most diesel-intensive activities in agricultural systems, especially in highly mechanized farms that rely on heavy machinery.

To address the energy challenges, alternatives to fossil-fuel-based machinery have been explored. \citet{lajunen_simulation-based_2024} analyzed the potential of hybrid and electric powertrains for agricultural tractors. Their findings show that while fully electric tractors are more energy-efficient, their limited autonomy poses challenges, especially in large-scale farming operations. Fuel cell hybrid tractors, on the other hand, provide a viable solution for long-duration, energy-intensive tasks, although the infrastructure for hydrogen refueling remains limited in agricultural regions.

\subsection{Global Robotics and Energy Consumption}

Energy consumption in robotic systems depends on various factors, including robot type, task complexity, and environmental conditions like temperature and dust \cite{soori_optimization_2023}. To mitigate these inefficiencies, energy-saving methods have been developed, that can be categorized into hardware-based and software-based strategies.

\subsubsection{Hardware optimization}

Hardware optimization involves selecting energy-efficient mechatronic systems that meet application requirements without excess. Enhancements include reducing the weight of robotic components by using lighter materials to decrease the energy required for movement \cite{soori_optimization_2023}. Another approach is to improve motor efficiency by minimizing rotor and stator losses, which contribute to energy waste \cite{saidur_review_2010}. The integration of energy-storing devices, such as flywheels and capacitors, enables energy recovery during operations. Kinetic Energy Recovery Systems (KERS), for example, store energy during deceleration and release it during subsequent movements, reducing energy loss through braking \cite{carabin_review_2017}. Additionally, regular maintenance helps maintain consistent energy optimization \cite{soori_optimization_2023}.

\subsubsection{Software optimization}

Beyond hardware, software optimizations offer another avenue for energy efficiency in robotic systems. These methods, focused on control algorithms and operational strategies, aim to minimize energy consumption through intelligent planning. It includes two key approaches: trajectory optimization and operation scheduling.

Trajectory optimization plays an important role in reducing energy consumption. For instance, optimizing the path of a robotic arm can smooth its movements and reduce energy consumption by up to 20\% by eliminating unnecessary motions \cite{soori_optimization_2023}. The Point-to-Point trajectory method considers each segment of movement as a sequence of two points without considering the optimization for subsequent steps, leading to energy savings of up to 12.5\%. In contrast, the Multi-Point method evaluates all points simultaneously, which, though more complex, allows for greater energy savings of up to 30\% through the use of optimization algorithms and models \cite{carabin_review_2017}.

Operation scheduling is another approach to improve energy efficiency by reducing idle time and minimizing queues. One technique, time scaling, adjusts the robot's velocity. Instead of focusing solely on task completion speed, this method reduces idle time by accelerating slower movements when possible and decelerating faster ones to prevent unnecessary energy consumption \cite{carabin_review_2017}. Sequence scheduling further improves energy efficiency by optimizing the order of operations, minimizing the energy impact of acceleration and deceleration. During idle periods, both velocity and acceleration are reduced, leading to additional energy savings \cite{soori_optimization_2023, carabin_review_2017}.

\subsubsection{Mixed approaches}

Mixed approaches combine both hardware and software strategies to maximize energy efficiency. By integrating energy recovery systems like KERS with optimized motion planning and control algorithms, these approaches provide a strategy for reducing the overall energy consumption of robotic systems \cite{carabin_review_2017}.

\subsection{Agricultural robotics and energy consumption}

Although energy-saving techniques have been explored for agricultural machines and general robotic systems, understanding the specific factors that influence energy consumption in robots designed for agricultural tasks is still in its early stages.

In agricultural robotics, locomotion accounts for a larger portion of energy use compared to other components such as computation and sensors. For example, in low-friction scenarios where a mobile robot is in motion, locomotion represents 50\% of the total energy consumption, while computation and sensors contribute 33\% and 11\%, respectively \cite{wu_review_2023}. In agricultural tasks, where friction is considerably higher \cite{otsu_energy-aware_2016}, the impact of locomotion on energy consumption increases further. As a result, the energy consumption of agricultural robots is closely influenced by factors such as friction, steering angle during turns, load weight, terrain type, ground slope, and the location of the center of gravity \cite{wu_review_2023}.

Weight load has a notable impact on energy consumption for mobile robots, with studies demonstrating a linear relationship between the two variables \cite{wu_review_2023, loukatos_power_2024}. Similarly, the turning radius of the robot’s trajectory plays a critical role in determining energy use, with larger turning radii shown to significantly reduce energy consumption \cite{hizatate_work_2023}. The center of mass also influences energy efficiency. When the payload is positioned on the outer edge during a turn, it acts as a counterweight, optimizing energy use, while placing it on the inner edge tends to increase energy consumption \cite{wu_review_2023}.

Simulations by \citet{kivekas_effect_2024}, involving a robot pulling a harrow on deformable terrain, revealed how energy losses are distributed between the robot’s tires and the harrow. At lower speeds, tire friction is the primary source of energy loss. However, as speed and the depth of the harrow’s tines increase, friction from the tool grows significantly, possibly matching the energy losses from the robot’s tires. In addition, \citet{valero_assessment_2019} demonstrated that at low speeds, energy consumption follows a linear pattern, but at higher velocities, energy usage increases disproportionately due to rising friction. Braking is another factor contributing to energy dissipation, with a near-linear relationship observed between energy consumption and dissipation during braking \cite{valero_assessment_2019}. To conserve energy, designing trajectories that maintain the robot’s inertia and minimize braking is essential.

Terrain type also plays a significant role in energy consumption, with sandy terrain consuming more energy than grassy or rough fields, which in turn consume slightly more than asphalt \cite{otsu_energy-aware_2016, loukatos_power_2024}. Additionally, several experiments have shown that the slope of the terrain affects energy consumption \cite{wu_review_2023, otsu_energy-aware_2016, hizatate_work_2023}. Optimizing trajectories to take advantage of favorable slopes can lead to energy savings, as surface inclination and speed have a greater impact on energy consumption than rough terrain alone \cite{loukatos_power_2024}. \autoref{tab:SOTA_parameters} presents a summary of the impacting factors that were described in this section.

\begin{table}[t!]
    \centering
    \caption{Factors influencing energy consumption sorted by their impact.}
    \renewcommand{\arraystretch}{1.5}
    \begin{tabularx}{\linewidth}{XXX}
        \toprule
        High impact & Medium impact & Moderate impact \\ \midrule
        Weigh load, Ground slope, Robot speed, Braking frequency, Task performed &%
        Center of mass location, Terrain type (Sandy) &%
        Turning radius, Terrain type (except sandy) \\
        \bottomrule
    \end{tabularx}
    \label{tab:SOTA_parameters}
\end{table}

\section{empirical analysis of energy consumption}
\label{sec:empirical_analysis}
\subsection{Experimental setup}

\begin{figure}[ht]
    \centering
    \includegraphics[trim=6cm 3cm 6cm 0cm, clip, width=\linewidth]{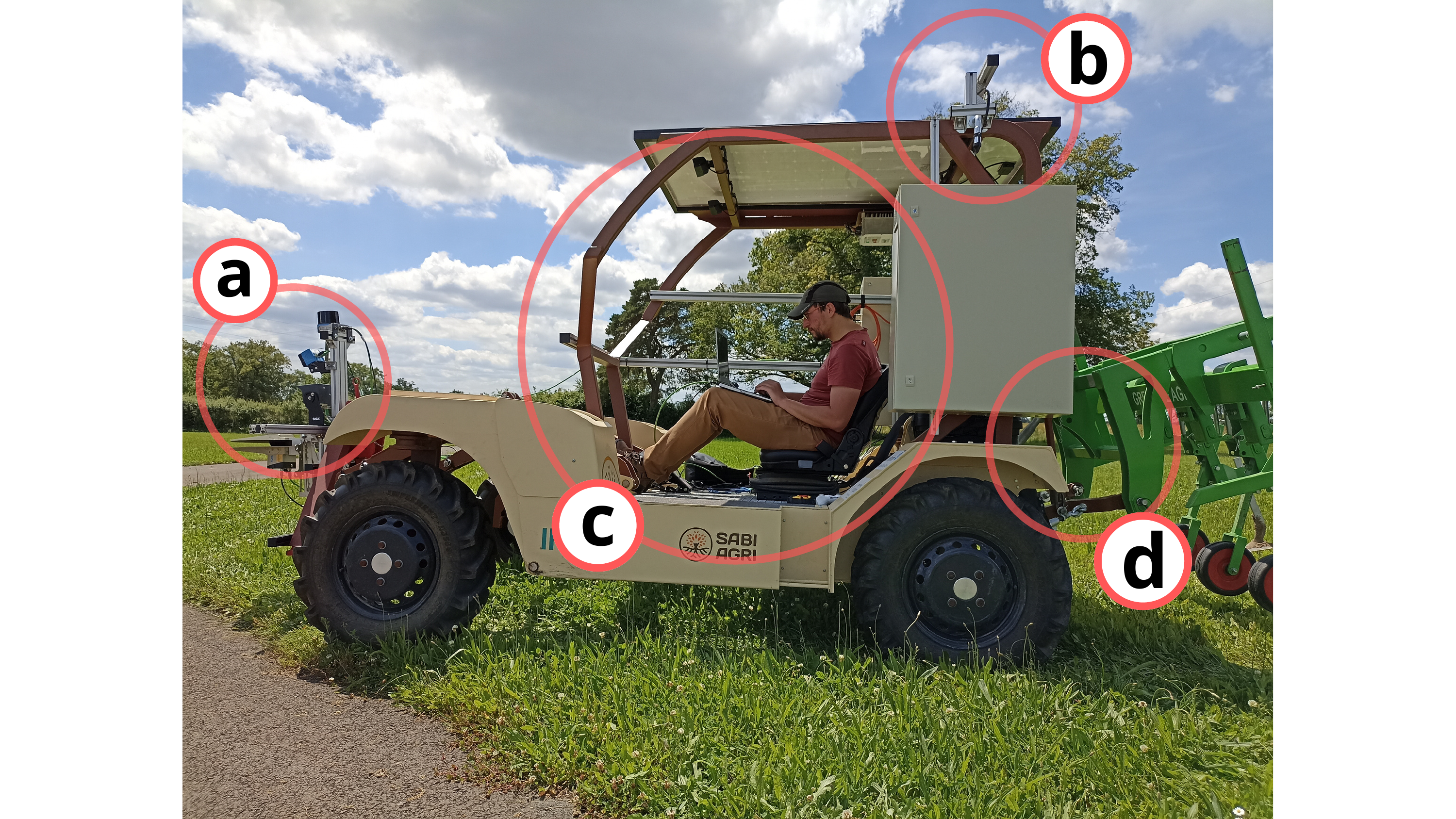}
    \caption{Automated tractor used in the experiment. (a) Lidar sensor; (b) GPS sensor; (c) driver location for possible supervision; (d) Attach point for agricultural tools. Note that although a person was present in the vehicle for clear safety purposes, the robot was fully autonomous during the experiments. }
    \label{fig:SabiAgri_Cyrille_le_BG}
\end{figure}

The experiments were conducted using an experimental robot, depicted in Figure \ref{fig:SabiAgri_Cyrille_le_BG}. 
The \textit{POM SabiAgri} is an automated ground robot specifically designed for agricultural tasks.
The robot weights \SI{1200}{\kg}, with a width of \SI{1.57}{\m} and length of \SI{2}{\m}.
The robot is equipped with two main sensors that allow its functionality and autonomous behavior. 
It features a lidar sensor, which is used for environmental scanning and obstacle detection. 
Additionally, the robot is fitted with a GNSS sensor that provides precise positioning data. This sensor allows the robot to register its position on datasets and to follow pre-defined GPS trajectories.

\begin{figure}[htbp]
    \centering
    \begin{subfigure}{0.245\linewidth}
        \includegraphics[width=\linewidth]{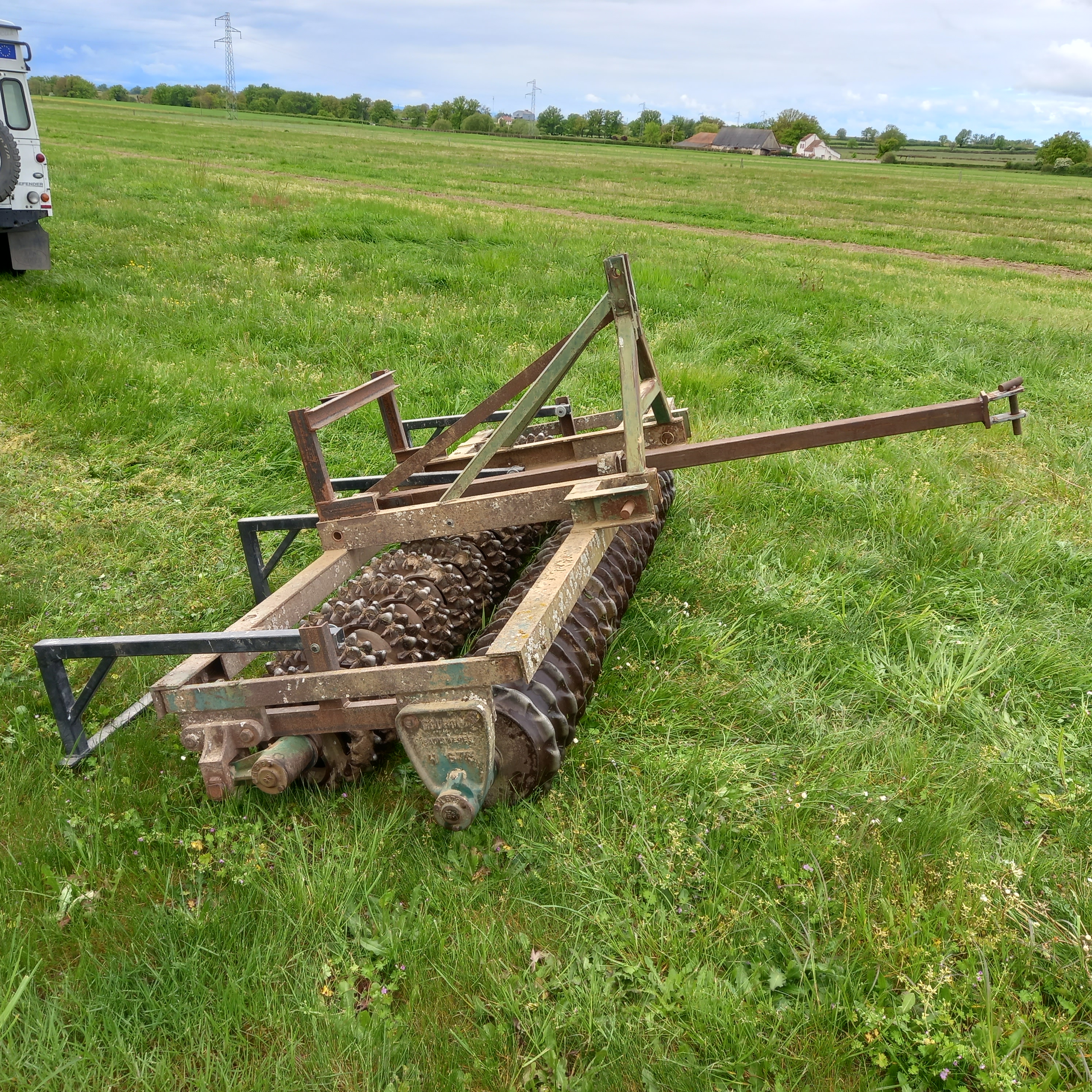}
        \caption{}
        \label{fig:roll}
    \end{subfigure}%
    \begin{subfigure}{0.245\linewidth}
        \includegraphics[width=\linewidth, angle=-90]{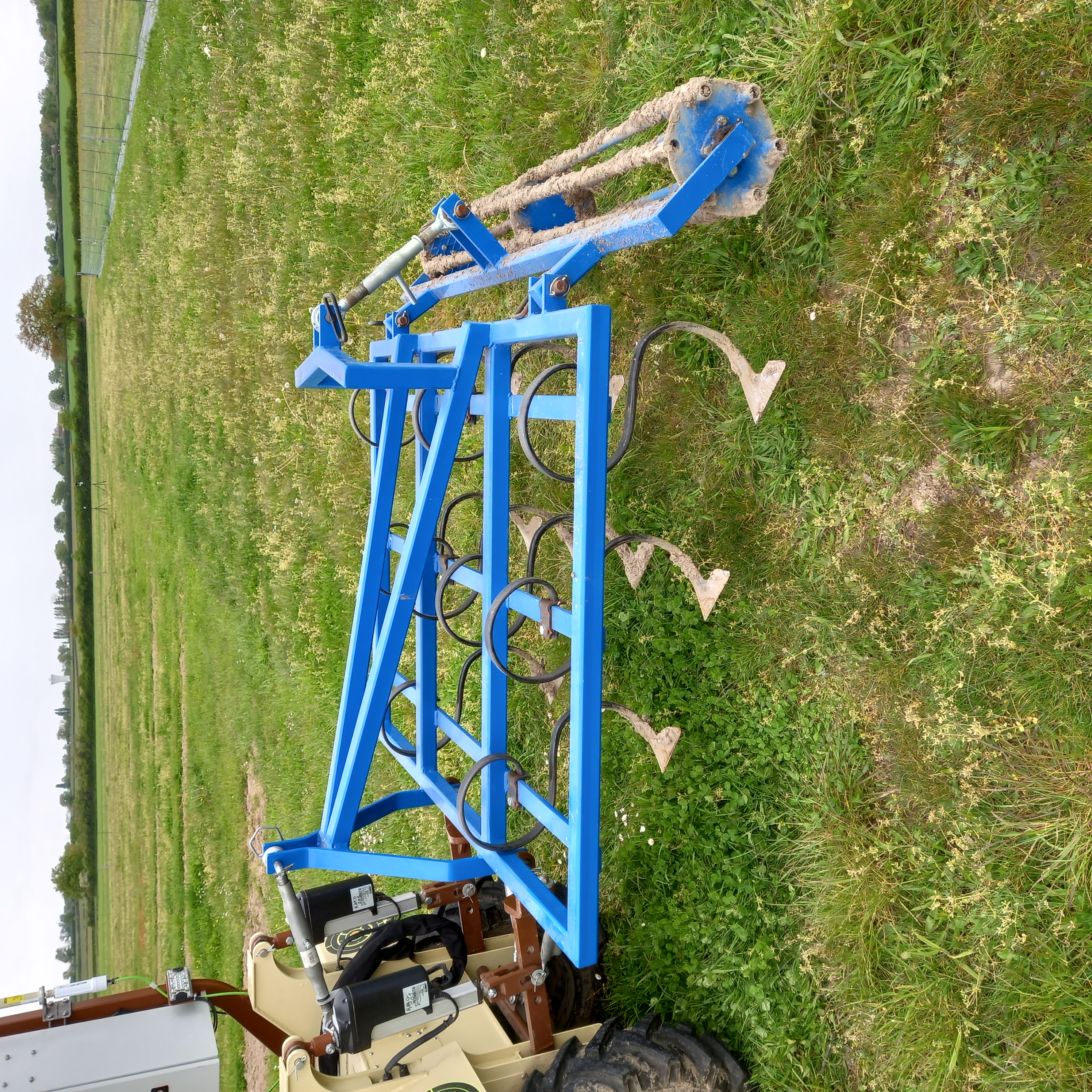}
        \caption{}
        \label{fig:vibro}
    \end{subfigure}%
    \begin{subfigure}{0.245\linewidth}
        \includegraphics[trim=14cm 0cm 17cm 0cm, clip, width=\linewidth]{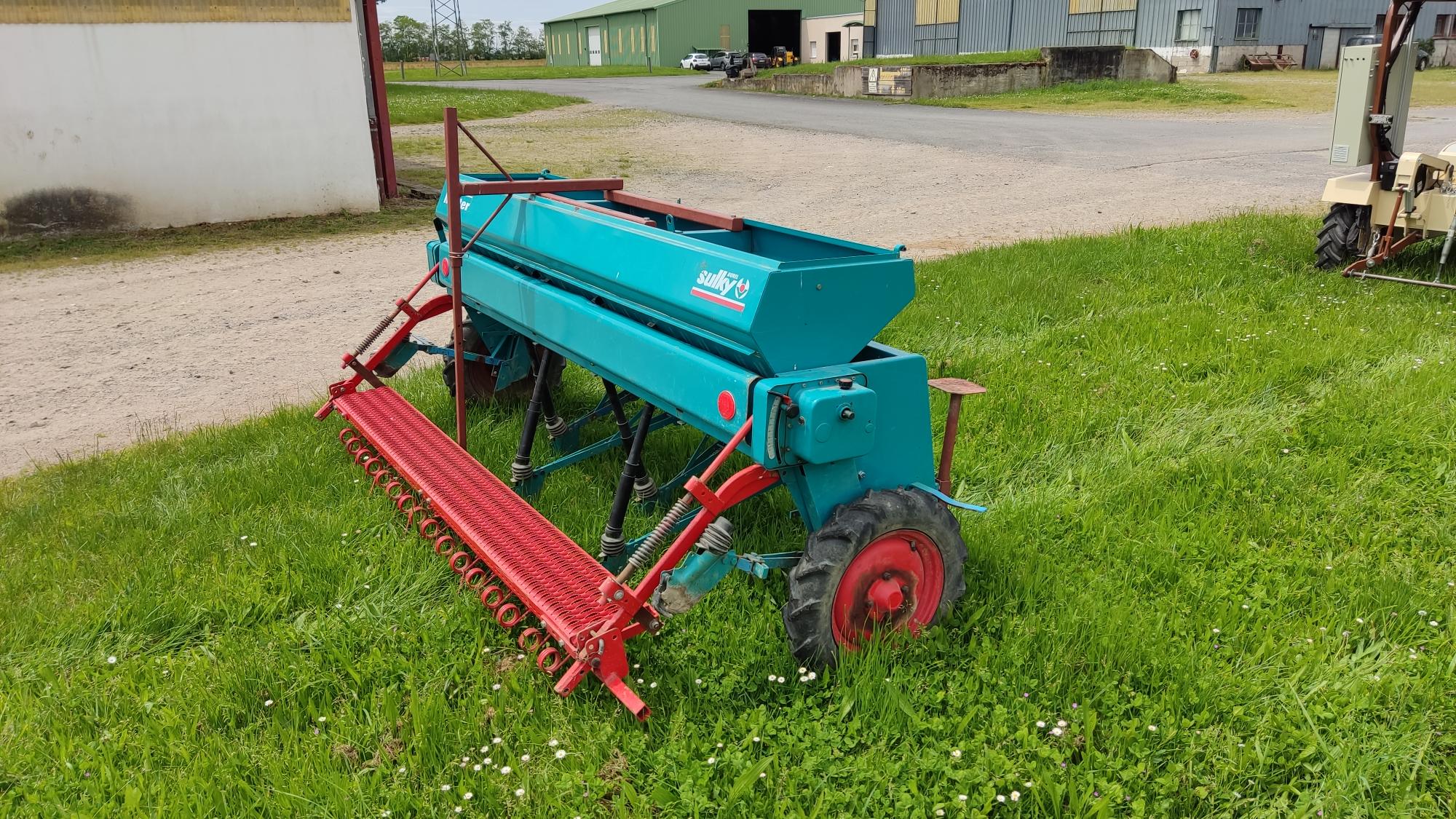}
        \caption{}
        \label{fig:seeder}
    \end{subfigure}%
    \begin{subfigure}{0.245\linewidth}
        \includegraphics[width=\linewidth]{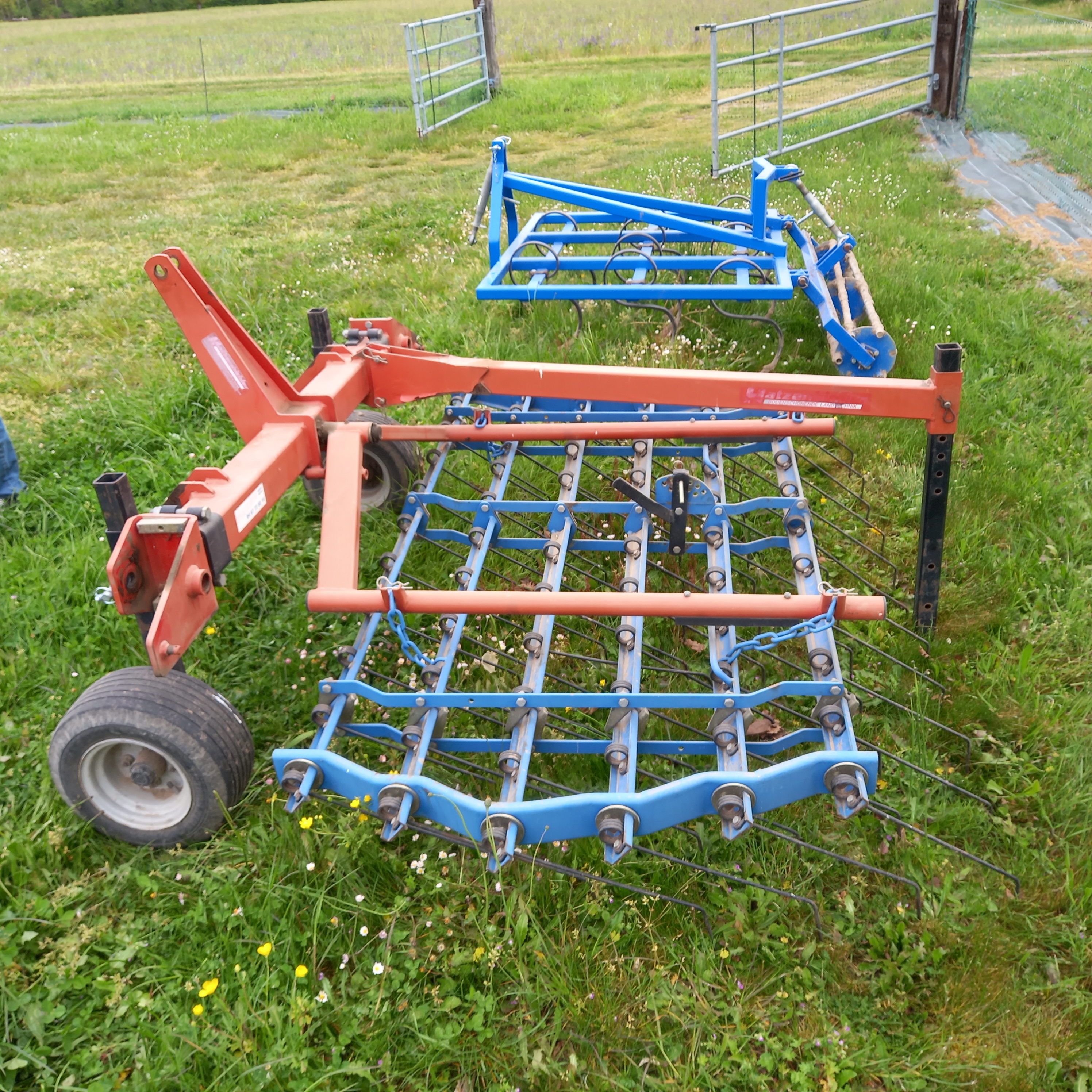}
        \caption{}
        \label{fig:harrow}
    \end{subfigure}%
    \caption{Agricultural tools used during the experiments: (a) Crosskill Roller; (b) Vibro-cultivator; (c) Seeder; (d) Harrow}
    \label{fig:tools}
\end{figure}

During the experiments, the robot performed tasks both with and without tools. The tools were fixed to the attached point at the back of the robot, as showed in Figure \ref{fig:SabiAgri_Cyrille_le_BG}. Four different tools were employed during the data collect, detailed in Figure \ref{fig:tools}:

The \textbf{crosskill roller}, illustrated in Figure \ref{fig:roll}, is primarily used for soil preparation. It is the most heavy tool used for the experiments with a weight of 1200 kg. It is effective in breaking up clods of earth post-plowing to create a finer and more even soil surface. This tool lightly compacts the topsoil, which aids in moisture retention and improves seed-to-soil contact, ultimately enhancing germination. Additionally, it helps level the soil and prepares an ideal seedbed, contributing to better crop growth.

The \textbf{vibro-cultivator}, shown in Figure \ref{fig:vibro}, uses vibrating tines to break up the soil, mix in crop residues, and eliminate weeds. Its weight is 192 kg, it is the lighter tool of the experimental sessions. This tool is designed to loosen and aerate the soil without completely overturning it, thus preserving soil structure and promoting healthy soil conditions.

The \textbf{seeder}, depicted in Figure \ref{fig:seeder}, is used to plant seeds accurately at the correct depth and spacing. This ensures even seed distribution across the field, improves planting efficiency, reduces seed wastage, and supports uniform crop emergence. Its weight is 450 kg.

The \textbf{harrow}, presented in Figure \ref{fig:harrow}, is used to break up clumps of soil, smooth the surface, and remove weeds. It also integrates crop residues into the soil and improves the seedbed for planting by creating a finer, more even surface. This tool is typically used after plowing to refine and level the soil before seeding.

The experiments were conducted at the National Research Institute for Agriculture, Food and the Environment (INRAE) experimental site in Montoldre, France.
These sites provided diverse environments, including multiple fields, grassy areas, and asphalt roads, which allowed for comprehensive testing of the robotic systems under various conditions.

\begin{figure}[htbp]
    \centering
    \includegraphics[trim=0.8cm 0cm 0.5cm 0cm, clip, width=\linewidth]{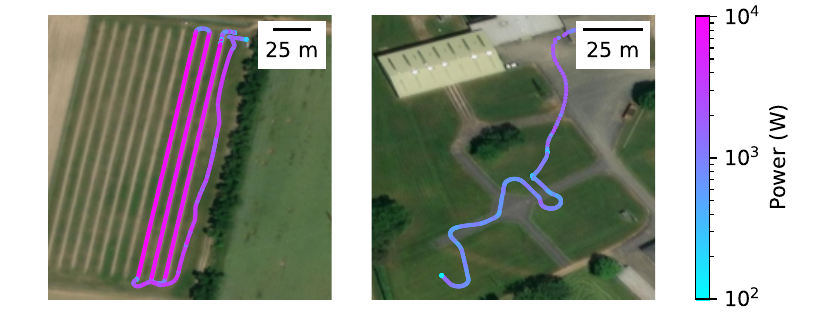}
    \caption{Example of trajectories performed by the agricultural robot, colored by the consumed energy. Left: The robot performed a task in a field while using the vibro-cultivator; Right: The robot is going back to the farm with no tool equipped.}
    \label{fig:traj_example}
\end{figure}

The data used in this study were collected during experiments focused on on-field (harrowing, seeding...) and off-field tasks (return trip between the field and the storage location), rather than specifically targeting energy consumption analysis.
Throughout these experiments, the robot gathered various types of data along its trajectories. 
In total, 69 trajectories were evaluated, for more than \SI{22}{\km} of total traveled distance. 
The types of data collected included several parameters such as time, battery capacity, battery temperature, charge of the battery, current observed at the battery terminals, voltage at the battery terminals, power in watts, robot’s position, and speed.
To supplement these data, weather information from the nearest weather station was added for each experiment day. 
The soil type of the experimental sites' ground was also manually segmented via GNSS images, with distinctions made between asphalt, camps, clay courts, grass, and experimental zones. Furthermore, information about the tools equipped by the robot during each experiment were included.
\autoref{fig:traj_example} provides some examples of the trajectories made by the robot.

\subsection{Results and analysis}

In this section, we present a concise analysis of the energy consumption based on the recorded dataset.

The robot's tasks varied in duration, ranging from as little as one minute to up to twenty-two minutes, with an average task duration of seven minutes. 
The energy consumption during these operations was significant; tasks consumed up to \SI{1260}{\W\hour} during data collection. The average consumption of experiments is \SI{250}{\W\hour}, with the robot being able to consume a power up to \SI{12}{\kW}.

\begin{figure}[htbp]
    \centering
    \includegraphics[trim=0.1cm 0cm 0cm 0.4cm, clip, width=\linewidth]{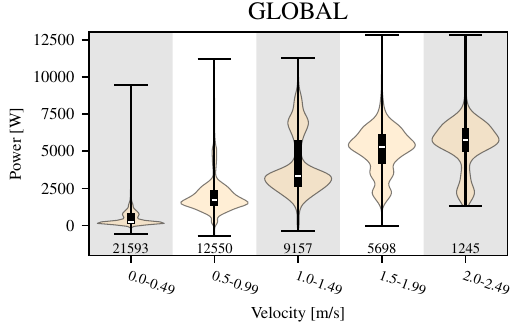}
    \caption{Violin plot relation between the speed level classes (0.5 m/s step) and the energy consumption. The black boxes represent the box plot with median in white, quartiles at 25\% and 75\% and extremum indicated by the whiskers. The underlying curves show the data distribution. The number below each violin represent the data count. }
    \label{fig:violin_speed_global}
\end{figure}

The velocity is a significant factor affecting energy consumption, regardless of the specific task being performed. 
As such, this parameter is used as the main factor in our analysis.
As illustrated in Figure \ref{fig:violin_speed_global}, there is a clear correlation between the overall power usage and the velocity of the robot. %
This observation is consistent with findings reported in previous studies, such as those in \cite{valero_assessment_2019}, \cite{kivekas_effect_2024}.
The distribution of energy consumption data around the median is generally homogeneous, except for the speed range of 1 to 1.49 m/s in Figure \ref{fig:violin_speed_global}. 
This specific speed level shows a slight bimodal distribution, which we hypothesize to be due to the fact that this speed level is attained in both scenarios where the robot is reaching the field (no tool) and actively working in the agricultural field (with a tool).

Additionally, there are few data points indicating negative power values for speeds between 0 and 1.5 m/s. 
This anomaly is attributed to instances where the robot lost connection and entered free wheel mode.
During these occurrences, the robot's inertia caused a slight recharge of the battery. 
Although this phenomenon is not desired, it presents an opportunity to potentially reduce energy losses by implementing a recovery system, as mentioned in \cite{saidur_review_2010}.

\begin{figure*}[thbp]
    \centering
    \includegraphics[trim=0.07cm 0cm 0cm 0cm, clip, width=\linewidth]{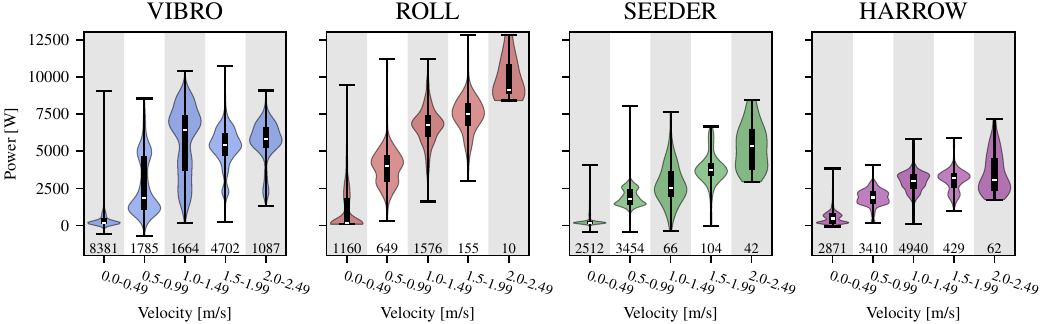}
    \caption{Energy consumption comparison between task performed and speed level associated. The black boxes represent the box plot with median in white, quartiles at 25\% and 75\% and extremum indicated by the whiskers. The underlying curves show the data distribution. The number below each violin represent the data count.}
    \label{fig:violin_speed_tool}
\end{figure*}

Furthermore, the type of task performed by the robot is a highly influential parameter on energy consumption. 
As shown in Figure \ref{fig:violin_speed_tool}, some tasks consume significantly more energy than others. 
It is obvious that increasing the speed has a more pronounced impact on energy consumption, especially when using tools that work at greater depths or that have a bigger weight, implying higher friction losses that are the main source of energy losses as illustrated in \cite{wu_review_2023}. 
This analysis shows that the robot should adjust the speed level depending on the task being performed. 
For instance, using the harrow at high speed can save time without drastically affecting energy consumption, while operating the crosskill roller at a lower speed helps to minimize energy usage. 
This approach of balancing speed and energy efficiency is a form of operations scheduling, as discussed by \citet{carabin_review_2017}.
This ascertainment is in line with the observations in \cite{kivekas_effect_2024}.
This highlights the possibility to lower the speed of execution for the more energy consuming tasks (e.g., crosskill roller), and to save time by increasing the speed level on the less energy consuming ones (e.g., harrow).

\section{Discussion}
\label{sec:discussion}

There is a clear need for more thorough energy consumption analysis within the field of agricultural robotics, as this topic remains little explored despite its significance for agroecological practices. 
This study provides a small overview of the subject, intending to highlight the importance of this topic in future robotic developments.
However, this type of analysis also presents certain challenges and must be considered as a starting point for energy analysis. The limited data availability prevents a comprehensive energy analysis, and the compilation nature of the data can lead to over-interpretation of the results graphs if the data is not sufficiently representative. A valid analysis requires a substantial amount of data to accurately reflect trends and correlations, which, in this study, are appreciated visually from the graphs but cannot be mathematically expressed. Additionally, the main parameter under evaluation, energy consumption, is influenced by a multitude of intrinsically linked factors, and the violin plot method used in this analysis cannot fully capture the complex interactions between these variables.
Furthermore, there are gaps in the data that hinder a complete understanding of the sources of energy losses. 
For instance, precise data about the environment, such as the humidity level of the ground, could bring precious additional insight about the robot's energy consumption and how to optimize it.

Indeed, the data collected during this study was not specifically intended for an empirical analysis of energy consumption, meaning that some aspects could have been optimized for more accurate results. In an outdoor agricultural environment, the parameters under study are often intrinsically linked, and the diversity of the collected data was too broad to yield concrete results across all variables. To improve future studies, it would be advisable to establish a protocol that allows for the examination of each parameter individually, ensuring the collection of the most relevant data.
Nevertheless, real-world data from agricultural tasks provides critical insights that may not be fully captured by benchmarks conducted in controlled environments. 
Real-world scenarios encompass the complexity and variability of actual farming conditions, which controlled settings may not adequately replicate. 
This makes field data essential for developing robust and adaptable robotic systems capable of operating effectively under diverse agricultural conditions.
Real experiments can elucidate which parameters are most significant for energy consumption. 
Ultimately, the focus should be on energy consumption during practical agricultural tasks rather than in idealized benchmarks. 
Benchmarks may highlight theoretical scenarios that are impractical or overlook the complex interactions among variables present in real-world conditions.

\section{Conclusion}
\label{sec:conclusion}

This study focused on the energy consumption of an automated robot during various agricultural tasks. 
Our experiments highlighted the significant role that speed and the type of task play in determining energy consumption. 
Greater velocities generally increased energy consumption, particularly when using heavy tools such as the crosskill roller. Conversely, lighter tools like the harrow allowed for higher speeds with relatively lower energy costs. These findings suggest that optimizing the speed and task allocation of agricultural robots can lead to significant energy savings.
We also identified the importance of task-specific strategies for energy optimization. For instance, reducing speed during energy-intensive tasks and increasing it during less demanding operations can enhance overall efficiency. This approach aligns with the concept of operation scheduling, which aims to balance time and energy use effectively.

Despite the insights gained, our analysis faced limitations due to the diverse and intrinsically linked nature of the experimental parameters. The variability in conditions such as weather, tool usage, and speed levels made it challenging to isolate specific factors influencing energy consumption. 

Future work will involve establishing a more structured protocol for future experiments. 
This should involve isolating and examining each parameter individually, collecting extensive and detailed data, and ensuring precise observations of the robot's state during tasks. Additionally, incorporating more environmental factors will provide a better comprehensive understanding of energy use.

\printbibliography

\end{document}